\documentclass{article}

\usepackage{microtype}
\usepackage{graphicx}
\usepackage{booktabs}
\usepackage{hyperref}
\usepackage{enumitem}
\usepackage{amsmath,amssymb}
\usepackage{tikz}
\usetikzlibrary{calc,positioning,arrows.meta}
\usepackage{float}
\usepackage{algorithm}
\usepackage{algorithmic}
\usepackage{caption}
\usepackage{needspace}

\usepackage[accepted]{icml2026}

\makeatletter
\renewcommand\paragraph{\@startsection{paragraph}{4}{\z@}%
  {0.6ex \@plus 0.2ex \@minus 0.1ex}%
  {-0.5em}%
  {\normalfont\normalsize\bfseries}}
\makeatother

\setlist{topsep=2pt,itemsep=1pt,parsep=0pt,partopsep=1pt}

\icmltitlerunning{When Retrieval Fails Before It Begins}

\makeatletter
\renewcommand{\Notice@String}{Accepted at the ICML 2026 Workshop on \textit{Failure Modes of Agentic AI} (FAGEN@ICML 2026). Non-archival; the authors retain all rights.}
\makeatother

\begin{document}
\raggedbottom

\twocolumn[
  \icmltitle{When Retrieval Fails Before It Begins:\\
    Structurally Indirect Prerequisite Eviction\\
    as a Retention Failure in Agentic Memory}

  \begin{icmlauthorlist}
    \icmlauthor{Minkyu Song}{yonsei}
  \end{icmlauthorlist}

  \icmlaffiliation{yonsei}{Yonsei University, Seoul, South Korea}
  \icmlcorrespondingauthor{Minkyu Song}{smkgenesis@yonsei.ac.kr}

  \icmlkeywords{LLM agents, memory, retention, graph, failure modes}

  \vskip 0.3in
]

\printAffiliationsAndNotice{}

\begin{abstract}
Agentic memory under a fixed budget involves two stages:
\emph{retention} and \emph{retrieval}. Existing retrieval-centered
paradigms implicitly assume
necessary evidence survives eviction, but we challenge this by isolating a
pre-retrieval failure mode: \emph{structurally indirect prerequisite
eviction}, in which upstream blocks weakly aligned with the query are
discarded under budget pressure. We provide an operational
definition of this failure, a reproducible deterministic benchmark, and
per-seed trace diagnostics. Finally, we evaluate \emph{Dependency-aware
Semantic Garbage Collection} (DSGC), a one-hop graph-aware rule. In our
main suite, DSGC improves full-chain retention from $0.03$ to $0.90$
under a lexical encoder and from $0.23$ to $1.00$ under a sentence
encoder. Robustness checks then identify the budget and scaling regimes
where the one-hop rule holds or degrades. Our released pipeline and
failure postmortem support mechanistic analysis of retention before
retrieval as a distinct failure boundary.
\end{abstract}

\section{Introduction}

Long-horizon LLM agents face two logically distinct decisions about past
context under a fixed budget: \emph{retention} (which blocks survive
eviction) and \emph{retrieval} (which surviving blocks are surfaced for
the current query). Existing work has primarily emphasized retrieval---how
to rank or expand evidence once a store is available
\cite{rag2020,realm2020,retro2022,atlas2023,graphrag2024,hipporag2024}.
We study an earlier failure boundary: the store may already be incomplete
because a necessary block was evicted before retrieval begins.

Common memory-management heuristics include keeping recent tokens, scoring
blocks by similarity to the current query, or paging memories in and out
heuristically \cite{memgpt2023,scm2023,generative_agents2023}. These
policies are effective when the needed evidence is recent or lexically
aligned with the query. However, they can fail in a specific way when the
answer depends on an upstream block that is weakly aligned with the query
but remains structurally required. We call such a block a
\emph{structurally indirect prerequisite}, and we make this precise in
Section~\ref{sec:def}. Prerequisite chains of this kind arise naturally
wherever an upstream fact determines the actionability of a downstream decision. Once a structurally indirect prerequisite has been
evicted under budget pressure, no downstream retrieval procedure can
recover it from the store. Retention is therefore
logically prior to retrieval in the failure chain: retrieval ranks what
remains; retention decides what remains.

This paper treats that retention-stage failure as a first-class object of
study. Concretely, we
\begin{enumerate}[leftmargin=1.3em,itemsep=1pt,topsep=2pt]
  \item give an \emph{operational definition} that distinguishes the
    failure from downstream retrieval failure, non-structural forgetting,
    and reasoning failure (Section~\ref{sec:def});
  \item construct a deterministic benchmark as a \emph{reproducible
    trigger} with fixed seeds, retrieval-friendly controls, and
    dependency-sensitive targets (Section~\ref{sec:bench});
  \item provide \emph{trace diagnostics} that decompose each failing
    seed into a displaced block, a displacing competitor, and a score
    margin (Section~\ref{sec:traces});
  \item evaluate \emph{Dependency-aware Semantic Garbage Collection}
    (DSGC), a one-hop graph-aware retention rule that protects
    prerequisites via structural reachability, and characterize the
    regimes where it holds and breaks
    (Sections~\ref{sec:method}, \ref{sec:main}--\ref{sec:robust}).
\end{enumerate}
Together, these components establish a controlled base case: with
supplied dependency edges, retention can be studied as a structural
reachability problem before retrieval begins, separately from graph
induction and downstream retrieval.

\subsection{Positioning}

We position the problem earlier in the agentic-memory pipeline than
retrieval: before evidence can be ranked or expanded, it must first
survive eviction. We contrast this setting with three established
paradigms and draw our primary inspiration from a fourth.

\paragraph{Retrieval and graph-augmented retrieval.} Standard
retrieval-augmented systems \cite{rag2020,realm2020,retro2022,atlas2023}
assume the relevant evidence is still present in the store at query time.
Graph-augmented retrieval methods such as GraphRAG and HippoRAG
\cite{graphrag2024,hipporag2024} use graph structures to expand or
re-rank evidence over a store that is already assumed to be complete. They
focus on downstream utilization, not upstream preservation.

\paragraph{Agentic memory systems.} Memory systems including MemGPT, SCM,
Generative Agents, Reflexion, Voyager, and LATS
\cite{memgpt2023,scm2023,generative_agents2023,reflexion2023,voyager2023,lats2024}
demonstrate that external memory management is critical for long-horizon
tasks, but their eviction policies are usually not designed for the case
studied here: preserving weakly query-aligned prerequisites under budget
pressure. Similarly, context-compression methods
\cite{gisting2023,longllmlingua2024} change how content is represented
inside the budget rather than deciding which semantic units survive
eviction; unlike block-level eviction, compression preserves a trace
of every block and leaves inter-block structural relations intact.

\paragraph{Systems-level garbage collection.} We draw framing inspiration
from tracing garbage collection \cite{mccarthy1960recursive,jones2011garbage},
and the name \emph{Dependency-aware Semantic Garbage Collection} reflects
this analogy directly. In tracing GC, a root set of live objects
propagates reachability to all objects they reference; unreachable
objects are collected (evicted). DSGC mirrors this structure at the
block level: query-facing blocks with high semantic relevance serve as
the root set, liveness propagates one hop along declared prerequisite
edges to structurally necessary predecessors, and blocks that receive
neither direct semantic support nor structural coverage are treated as
collectable. The \emph{semantic} qualifier marks that the root set is
selected by query similarity rather than by syntactic pointers; the
\emph{dependency-aware} qualifier marks that liveness is not purely
similarity-based but propagates through the prerequisite graph---the
structural extension that vanilla semantic selection omits. This framing
treats agentic context eviction as a reachability-style retention problem
rather than only a semantic ranking problem.

\section{Operational Definition}
\label{sec:def}

\subsection{Setting and Terminology}

A \emph{memory store} contains $N$ blocks
$\mathcal{B}=\{b_1,\ldots,b_N\}$. A \emph{block} is a semantically atomic unit---a single,
indivisible proposition---that is either retained or evicted in full;
it carries a token cost $s_i$. In our experiments, $s_i$ is a
deterministic whitespace word count,
$\max(|\mathrm{text.split()}|,1)$, used consistently by all reported
budgets and reproduction scripts, ensuring cross-tokenizer
reproducibility. The benchmark also supplies a directed prerequisite graph $E$.
An edge $(j,i)\in E$ means block $b_j$ depends on block $b_i$, so $b_i$
is an \emph{immediate prerequisite} of $b_j$. The benchmark supplies
prerequisite edges as ground truth, allowing the experiment to test
retention independently of graph induction.

Given a query $q$ and a token budget $C<\sum_i s_i$, each retention policy
induces a ranking over blocks. The retained subset
$S\subseteq\mathcal{B}$ is then constructed greedily: blocks are added in
policy order until adding the next block would make
$\sum_{b_i\in S}s_i>C$. Retrieval downstream of retention is restricted
to $S$: any block not in $S$ is unavailable to downstream retrieval from
the active prompt-visible store for this query step. We use
$\mathrm{sim}_\phi(q,b_i)$ for the similarity score between query $q$ and
block $b_i$ under encoder $\phi$, and write $\mathrm{chain}(q)$ for the
generator-labeled set of blocks whose joint retention is necessary to
recover the benchmark answer to $q$.

\subsection{Failure Mode}

The pre-retrieval failure mode has two components: a block's structural
vulnerability (\emph{structurally indirect prerequisite}) and the eviction
event that realizes it (\emph{retention-stage failure}).

\paragraph{Definition 1 (Structurally indirect prerequisite).} A chain
block $b_i\in\mathrm{chain}(q)$ is \emph{structurally indirect with
respect to $q$ under encoder $\phi$} when there exists another chain
block $b_j$ that depends on it and that has higher query similarity:
\begin{equation}
\label{eq:indirect}
\exists\, b_j\in\mathrm{chain}(q):\;(j,i)\in E\;\wedge\;
\mathrm{sim}_\phi(q,b_i)<\mathrm{sim}_\phi(q,b_j).
\end{equation}
Intuitively, $b_i$ is a necessary chain block whose text is less
query-aligned than the downstream block $b_j$ it enables. Indirectness
is a property of the (block, query, encoder) tuple, not of any policy.

\paragraph{Definition 2 (Retention-stage failure).} A retention policy
exhibits a \emph{retention-stage failure} on $(q,C)$ when it produces a
selection $S$ that excludes a structurally indirect prerequisite while
retaining the more query-similar downstream block that depends on it.
The failure occurs before retrieval: once $b_i$ leaves the active
prompt-visible store, downstream ranking cannot recover it.

\paragraph{What this excludes.} Definition~2 excludes three
adjacent phenomena. (i) \emph{Retrieval failure}: the needed block is in
$S$ but the retriever ranks it below an unhelpful competitor. (ii)
\emph{Surface forgetting}: $b_i$ is dropped because it is simply older or
because no chain of any kind connects it to $q$. (iii) \emph{Reasoning
failure}: $S$ is complete but the model misuses it. We isolate the case
in which structural necessity and surface similarity disagree at the
eviction step.

\paragraph{Worked example.} Consider three blocks: $c_1$:~``Alice is
assigned to backup bundle M''; $c_2$:~``Bundle M maps to clearance N'';
$c_3$:~``Clearance N permits deletion of the encrypted backup.'' Edges
$(c_3,c_2)$ and $(c_2,c_1)$ encode the prerequisite chain; the query is
``May Alice delete the encrypted backup?'' Under a lexical encoder,
$\mathrm{sim}_\phi(q,c_3)$ is high (``delete'', ``backup''),
$\mathrm{sim}_\phi(q,c_1)$ is moderate (``Alice''), but
$\mathrm{sim}_\phi(q,c_2)$ is low because $c_2$ shares no surface terms
with the query. Here, $c_2$ satisfies Definition~1 with witness $b_j=c_3$:
it is a necessary chain block yet sits in a ``semantic valley'' between
two more relevant blocks. Under budget pressure, a similarity-only policy
identifies $c_2$ as the primary eviction candidate, destroying the logical
bridge between Alice and her permission; DSGC treats $c_3$ as a liveness
root and propagates survival support to $c_2$, preserving the chain.

\section{Method: One-Hop Dependency Propagation}
\label{sec:method}

We study a minimal one-hop graph-aware retention rule,
Dependency-aware Semantic Garbage Collection (DSGC). Rather than directly
searching for ``garbage'' to evict, DSGC propagates survival support from
highly relevant downstream blocks to their structurally necessary
prerequisites. The rule excludes learned components, multi-hop diffusion,
and per-step reranking, making the effect of one-hop dependency
propagation directly testable.

\begin{figure}[b]
\centering
\resizebox{0.95\columnwidth}{!}{%
\begin{tikzpicture}[
  block/.style={draw, rounded corners=3pt, minimum width=2.1cm, minimum height=0.7cm, align=center, line width=0.4pt, font=\small},
  query/.style={draw, rounded corners=3pt, fill=blue!6, minimum width=1.5cm, minimum height=0.7cm, align=center, line width=0.4pt, font=\small},
  note/.style={draw, rounded corners=3pt, fill=green!6, minimum width=2.2cm, minimum height=0.7cm, align=center, line width=0.4pt, font=\small},
  ann/.style={inner sep=1pt, font=\scriptsize, text=black!80},
  propann/.style={fill=white, inner sep=1pt, rounded corners=2pt, font=\scriptsize, text=green!50!black},
  arr/.style={-{Latex[length=2mm]}, thick},
  dep/.style={-{Latex[length=2mm]}, thick, dashed}
]
\node[query] (query) {Query};
\node[block, right=0.7cm of query] (c3) {Block $c_3$\\high relevance};
\node[block, below=0.85cm of c3] (c2) {Block $c_2$\\low relevance};
\node[block, below=0.85cm of c2] (c1) {Block $c_1$};
\node[note, right=0.9cm of c2] (score) {$I_i = r_i + \lambda_\pi \pi_i$};

\draw[arr] (query) -- (c3);
\draw[dep] (c3) -- (c2);
\draw[dep] (c2) -- (c1);
\draw[arr, green!50!black] (c3.east) -- (score.west);
\draw[arr, green!50!black] (score.west) -- ++(-0.45,0) |- (c2.east);
\node[ann] at ($(query.east)!0.5!(c3.west)+(0,-0.5)$) {relevance};
\node[ann, anchor=west] at ($(c3)!0.5!(c2)+(0.12,0)$) {dependency};
\node[ann, anchor=west] at ($(c2)!0.5!(c1)+(0.12,0)$) {beyond one-hop};
\node[propann] at ($(c3)!0.55!(score)+(0,0.3)$) {propagate $r_{c_3}$};
\end{tikzpicture}
}
\caption{The DSGC mechanism. A highly relevant downstream block ($c_3$)
elevates the retention score of its structurally necessary but semantically
indirect prerequisite ($c_2$) via the propagation term $\pi_i$. The
one-hop propagation targets retention, while multi-step inference remains
the role of the downstream reasoner. The policy halts after one hop;
vulnerabilities beyond this radius (e.g., $c_1$) constitute the residual
failure mode analyzed in Section~\ref{sec:traces}.}
\label{fig:mechanism}
\end{figure}
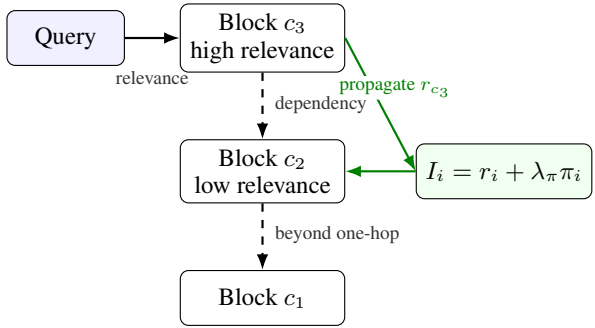

\paragraph{Scoring rule.} For query embedding $\mathbf{g}=\phi(q)$ and
block embedding $\mathbf{e}_i=\phi(b_i)$, raw relevance is normalized
over all $N$ blocks in the current store:
\[
r_i =
\frac{\exp(\mathbf{g}^{\top}\mathbf{e}_i/\tau)}
     {\sum_{k=1}^{N}\exp(\mathbf{g}^{\top}\mathbf{e}_k/\tau)},
\quad \tau = 0.25.
\]
We fix $\tau = 0.25$ before running the main suite.
The one-hop propagation term lifts relevance from a downstream block to
its immediate prerequisite:
\[
\pi_i = \sum_{j:(j,i)\in E} r_j.
\]
The \emph{DSGC score} combines the two additively:
\[
I_i = r_i + \lambda_\pi \pi_i,\qquad \lambda_\pi = 1.0.
\]
The main suite uses $\lambda_\pi = 1.0$ as the equal-weight base case:
one unit of downstream relevance contributes one unit of propagated
prerequisite support. Section~\ref{sec:robust} reports a
$\lambda_\pi$ sweep around this value. Selection is greedy in descending
$I_i$ until the additive token budget $C$ is exhausted; score ties preserve the
deterministic seeded block order used to instantiate the scenario. The
released policy implementation also supports a length-density variant; in
all reported experiments we set the density exponent to $0.0$, so score
differences reflect semantic relevance and propagated prerequisite support
rather than length normalization.

\paragraph{Three policies.} We compare three retention policies to cleanly
ablate the effect of graph propagation:
\begin{itemize}[leftmargin=1.3em,itemsep=1pt,topsep=2pt]
  \item \textbf{Similarity-only:} scores blocks by raw relevance $r_i$
        alone and greedily retains them until budget $C$ is exhausted.
        This is the standard semantic baseline that ignores graph edges.
  \item \textbf{No-graph DSGC:} runs the DSGC framework with
        $\lambda_\pi=0$ (so $I_i=r_i$); because $\exp$ is monotone,
        this preserves the exact ranking of the similarity-only baseline
        while executing within the unified DSGC code path.
  \item \textbf{DSGC:} the complete proposed rule with
        $\lambda_\pi=1.0$.
\end{itemize}
This three-way comparison separates code-path effects from the structural
propagation term: similarity-only and no-graph DSGC test ranking
equivalence, while no-graph DSGC versus DSGC tests the effect of
$\lambda_\pi\pi_i$.

\paragraph{Procedure.} The five-step implementation follows directly
from the scoring rule: encode all blocks and the query, compute softmax
relevance $r_i$, propagate one hop to form $\pi_i$, combine into
$I_i$, and greedily select under budget $C$.

\begin{algorithm}[H]
\begin{algorithmic}[1]
\REQUIRE blocks $\mathcal{B}=\{b_i\}$, query $q$, budget $C$, edges $E$, weight $\lambda_\pi$, encoder $\phi$
\STATE $\mathbf{g}\!\leftarrow\!\phi(q)$;\quad $\mathbf{e}_i\!\leftarrow\!\phi(b_i)$ \textbf{for all} $b_i\in\mathcal{B}$
\STATE $r_i \leftarrow \exp(\mathbf{g}^{\!\top}\mathbf{e}_i/\tau)\;/\;\textstyle\sum_k\exp(\mathbf{g}^{\!\top}\mathbf{e}_k/\tau)$\quad ($\tau=0.25$)
\STATE $\pi_i \leftarrow \sum_{j:\,(j,i)\in E} r_j$\quad \textbf{for all} $b_i\in\mathcal{B}$
\STATE $I_i \leftarrow r_i + \lambda_\pi\,\pi_i$\quad \textbf{for all} $b_i\in\mathcal{B}$
\STATE Sort $\mathcal{B}$ descending by $I_i$; greedily retain $b_i$ while $\textstyle\sum_{b_i\in S}s_i\leq C$
\ENSURE retained subset $S\subseteq\mathcal{B}$
\end{algorithmic}
\end{algorithm}

For $N$ blocks of embedding dimension $d$ and $|E|$ edges, scoring costs
$\mathcal{O}(Nd)$, propagation adds $\mathcal{O}(|E|)$, and selection is
$\mathcal{O}(N\log N)$.
This complexity profile characterizes the retention rule itself;
Section~\ref{sec:discussion} discusses the production-systems
implications.

\paragraph{Dependency graph.} The benchmark supplies prerequisite edges
directly, allowing us to test retention independently of graph induction.
This separation is useful because trace-rich agent settings
\cite{generative_agents2023,reflexion2023} suggest a plausible route to
future dependency induction, but do not remove it as a separate problem.
Our question here is therefore mechanistic: given dependency edges, how
much retention improvement can a minimal graph-aware propagation rule
produce?

\paragraph{Encoders.} We evaluate two encoders: an L2-normalized,
256-dimensional lexical encoder as a strict structural stress test for
surface-overlap vulnerabilities, and
\emph{all-MiniLM-L6-v2}
\cite{minilm2020,sentencebert2019}, producing 384-dimensional normalized
dense embeddings. The
lexical setting stresses surface-overlap failure directly; the dense
setting tests whether the same retention-stage failure persists under
stronger semantic embeddings. We report both because the failure persists
across encoder regimes, though the quantitative dynamics differ.

\section{Reproducible Trigger}
\label{sec:bench}

A reproducible trigger requires two ingredients: a test regime where
the failure mode reliably surfaces under budget pressure, and a control
regime where the structurally indirect prerequisite pattern
(Definition~1) is absent. Comparing these regimes lets us test whether
propagation helps specifically when structural necessity and query
similarity diverge.

\subsection{Controls and Targets}

\paragraph{Scenario layout.} Each scenario is constructed with a fixed
capacity of $20$ total blocks---small enough to keep tight budget regimes
informative yet large enough to admit real distractor competition. The
blocks fall into three distinct functional types:
\begin{itemize}[leftmargin=1.3em,itemsep=1pt,topsep=2pt]
  \item \textbf{Chain blocks:} The structurally necessary,
        generator-labeled blocks in $\mathrm{chain}(q)$ required to
        answer the query.
  \item \textbf{Honeypot blocks:} Semantic distractors (labeled
        \texttt{h1}, \texttt{h2}, etc.\ in traces) that share the
        query's decisive vocabulary but do not complete the reasoning
        chain and lack prerequisite edges into the chain. These are
        designed to reliably bait similarity-only retention policies.
  \item \textbf{Filler blocks:} Background noise (labeled \texttt{f1},
        \texttt{f2}, etc.) generated from neutral templates: short
        statements about disjoint entities, operations notes, or
        scheduling events that contain neither the answer relation nor
        the query's decisive vocabulary.
\end{itemize}
After the chain and honeypots are injected, the remaining positions in
the $20$-block context are occupied by filler blocks.

\paragraph{Budget multiplier.} Let
$L(q)=\sum_{b_i\in\mathrm{chain}(q)}s_i$ be the total token cost of the
necessary chain blocks. A \emph{budget multiplier} $M$ sets the retention
budget to $C=\lceil M\cdot L(q)\rceil$. We consider:
\begin{itemize}[leftmargin=1.3em,itemsep=1pt,topsep=2pt]
  \item $M=1.0$: critically tight.
  \item $M=2.0$: intermediate.
  \item $M=5.0$: loose.
\end{itemize}
We fix $M=2.0$ for the main suite because it provides enough aggregate
capacity for the chain to survive while preserving real competition at
the cutoff. At $M=1.0$, survival depends almost entirely on perfect
ranking; at $M=5.0$, most blocks survive and the policy distinction
largely disappears. Section~\ref{sec:robust} reports all three regimes.

\paragraph{Control template.} A scenario is a \emph{control} when its
necessary chain blocks share sufficient surface evidence with the query,
ensuring that a standard similarity-only policy can reliably retain the
chain. Controls test whether similarity-only retention succeeds when the
required blocks are query-aligned, and whether DSGC preserves this
behavior. We employ two control tasks:
\begin{itemize}[leftmargin=1.3em,itemsep=1pt,topsep=2pt]
  \item \textbf{Expense Control:} One block states a project budget,
        another sets an approval threshold, and the query asks whether
        a proposed expense is allowed.
  \item \textbf{Threshold Control:} The query names a specific
        thresholded entity and asks whether it crosses a defined
        numerical limit.
\end{itemize}
Both scenarios admit a similarity-only solution; a third candidate was
rejected for instability during calibration and is preserved in the
repository history. The accepted controls serve as ceiling checks:
degradation on these controls would indicate that propagation is harming
otherwise retrieval-friendly cases.

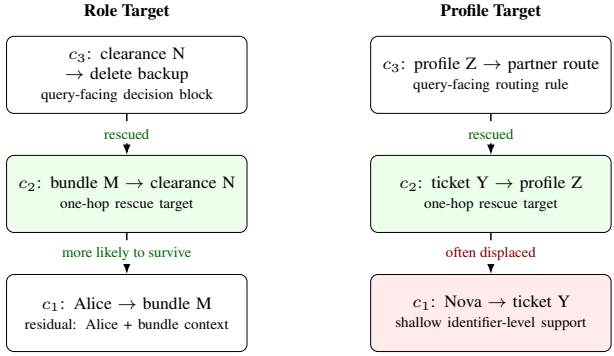
\begin{figure}[H]
\centering
\resizebox{0.97\columnwidth}{!}{%
\begin{tikzpicture}[
  box/.style={draw, rounded corners=3pt, text width=3.2cm,
              minimum height=1.1cm, align=center,
              line width=0.3pt, font=\scriptsize},
  qbox/.style={box, fill=green!8},
  nbox/.style={box, fill=white},
  wbox/.style={box, fill=red!8},
  midlbl/.style={font=\tiny, text=green!40!black,
                 fill=white, inner sep=1.5pt},
  midlblf/.style={font=\tiny, text=red!50!black,
                  fill=white, inner sep=1.5pt},
  dep/.style={-{Latex[length=1.5mm]}, semithick, dashed},
  ttl/.style={font=\scriptsize\bfseries},
]
\node[ttl] at (0,4.1) {Role Target};
\node[nbox] (Lc3) at (0,3.2) {$c_3$: clearance N $\to$ delete backup\\{\tiny query-facing decision block}};
\node[qbox] (Lc2) at (0,1.5) {$c_2$: bundle M $\to$ clearance N\\{\tiny one-hop rescue target}};
\node[nbox] (Lc1) at (0,-0.2) {$c_1$: Alice $\to$ bundle M\\{\tiny residual: Alice + bundle context}};
\draw[dep] (Lc3) -- (Lc2);
\draw[dep] (Lc2) -- (Lc1);
\node[midlbl] at (0,2.35) {rescued};
\node[midlbl] at (0,0.65) {more likely to survive};
\node[ttl] at (5.2,4.1) {Profile Target};
\node[nbox] (Rc3) at (5.2,3.2) {$c_3$: profile Z $\to$ partner route\\{\tiny query-facing routing rule}};
\node[qbox] (Rc2) at (5.2,1.5) {$c_2$: ticket Y $\to$ profile Z\\{\tiny one-hop rescue target}};
\node[wbox] (Rc1) at (5.2,-0.2) {$c_1$: Nova $\to$ ticket Y\\{\tiny shallow identifier-level support}};
\draw[dep] (Rc3) -- (Rc2);
\draw[dep] (Rc2) -- (Rc1);
\node[midlbl] at (5.2,2.35) {rescued};
\node[midlblf] at (5.2,0.65) {often displaced};
\end{tikzpicture}%
}
\caption{Differential survival of the multi-hop prerequisite ($c_1$).
Both target scenarios share the same structural retention architecture:
a query-facing downstream block ($c_3$) and an immediate prerequisite
($c_2$) successfully rescued via one-hop propagation. The critical
divergence occurs at the oldest block ($c_1$), which lies beyond the
explicit one-hop rescue radius. In the \textbf{Role Target} (left),
$c_1$ retains substantial residual lexical support (e.g., user identity
and task context), increasing its likelihood of survival. In the
\textbf{Profile Target} (right), $c_1$ possesses only shallow
identifier-level overlap and lacks the query's core routing vocabulary,
making it highly susceptible to displacement under budget pressure.
This contrast marks the boundary of the one-hop policy.}
\label{fig:targets}
\end{figure}

\paragraph{Target template.} A scenario is a \emph{target} when its
reasoning chain contains at least one structurally indirect prerequisite
under Definition~1. We evaluate two target architectures based on the
three-block chain ($c_1 \leftarrow c_2 \leftarrow c_3$):
\begin{itemize}[leftmargin=1.3em,itemsep=1pt,topsep=2pt]
  \item \textbf{Role Target} (moderate difficulty): The query asks
        whether Alice may delete a production backup. The chain consists
        of an older assignment block ($c_1$: Alice $\to$ bundle), an
        immediate prerequisite ($c_2$: bundle $\to$ clearance), and a
        query-facing decision rule ($c_3$: clearance $\to$ delete).
        Under DSGC, $c_3$ successfully rescues $c_2$ via one-hop
        propagation. The survival of the full chain then depends on
        whether $c_1$ retains enough residual lexical support (user
        identity and bundle context) to stay above the budget cutoff.
  \item \textbf{Profile Target} (high difficulty): Same three-stage
        architecture, but $c_1$ is stripped of decision-bearing routing
        vocabulary and retains only bare identifier-level support. $c_2$
        is still rescued by propagation, while $c_1$ is far more easily
        displaced under budget pressure (Figure~\ref{fig:targets}).
\end{itemize}

\paragraph{Honeypot strength.} Our data generator defines three honeypot
tier pools---\emph{easy}, \emph{medium}, and \emph{hard}. The main suite
uses the medium pool:
\begin{itemize}[leftmargin=1.3em,itemsep=1pt,topsep=2pt]
  \item \textbf{Easy:} too weak to challenge the cutoff.
  \item \textbf{Medium:} shares decisive query vocabulary but cannot
    complete the answer chain.
  \item \textbf{Hard:} risks creating confounded alternative paths.
\end{itemize}
Medium is the strongest adversarial setting that preserves a unique
ground-truth chain.

\paragraph{Synthetic templates.} The benchmark uses compact synthetic
templates---four templates with twenty blocks per scenario---for three
reasons:
\begin{enumerate}[leftmargin=1.3em,itemsep=1pt,topsep=2pt]
  \item \textbf{Ground-truth traceability.} Synthetic templates allow us
    to explicitly declare $\mathrm{chain}(q)$ as ground truth. This
    makes Definition~2 directly checkable: a failure decomposes into a
    specifically known displaced block, rather than manifesting as a
    generic incorrect answer with an opaque cause.
  \item \textbf{Controlled regime contrast.} Controls and targets share
    identical total block counts, budget rules, seed protocols, and
    encoder configurations. They differ only where the regimes require
    it: controls are retrieval-friendly, while targets instantiate
    structurally indirect prerequisites. The contrast is therefore
    between two explicitly declared regimes, isolating the structural
    variable of interest.
  \item \textbf{Falsifiability.} The synthetic setting permits us to
    lock the generation protocol and achieve bit-exact reproduction of
    the failure trigger.
\end{enumerate}

\subsection{Fixed Evaluation Protocol and Auditability}

The release preserves the calibration history, rejected variants, and
postmortem notes, making the final trigger auditable rather than only
reproducible. Each rejected variant is paired with the specific
calibration issue it exposed---ranging from insufficient distractor
strength to unstable seed behavior---so that subsequent benchmarks can
refine rather than rediscover the trigger.

\paragraph{Fixed evaluation protocol.} The final main suite locks all
parameters to prevent post-hoc tuning. The grid evaluates:
\begin{itemize}[leftmargin=1.3em,itemsep=1pt,topsep=2pt]
  \item \textbf{Templates:} 4 ($2$ controls, $2$ targets)
  \item \textbf{Policies:} 3 (similarity-only, no-graph DSGC, DSGC)
  \item \textbf{Encoders:} 2 (lexical, sentence)
  \item \textbf{Configuration:} $15$~seeds per condition,
    $20$~total blocks per scenario, medium-strength honeypots, budget
    multiplier $M=2.0$.
\end{itemize}
To ensure fair comparison, the protocol standardizes structural
complexity. \emph{Chain depth}---the total number of logically
sequential blocks required to answer the query---varies strictly by
template type: controls use a chain depth of~$2$ (with $4$~honeypots),
while targets use a chain depth of~$3$ (with $2$~honeypots). These
values are the smallest chain depths that separate the control and
target regimes cleanly. Depth~$3$ is the smallest chain in which one
block ($c_2$) sits exactly one hop from the query-facing block ($c_3$)
and one block ($c_1$) lies two hops away---the minimum configuration in
which the structural failure of Definition~1 is isolable without
confounding recency effects. Depth~$2$ controls therefore check whether
propagation preserves cases where direct query alignment is already
sufficient.

\paragraph{Evaluation metrics.} The primary metric is \textbf{Full Chain
Retention}: the binary success rate over seeded runs in which every
generator-labeled chain block survives the eviction budget. We also record
deterministic template answer accuracy, but in this suite it is nearly a
consequence of full-chain retention rather than an independent reasoning
metric. Means and standard deviations are reported over $15$~seeds.

\section{Main Results}
\label{sec:main}

\begin{table}[H]
\centering
\small
\begin{tabular}{@{}llll@{}}
\toprule
Encoder & Regime & Method & Retention \\
\midrule
Lexical  & Control & Sim.-only & 1.00 {\scriptsize$\pm$0.00} \\
Lexical  & Control & No-graph DSGC  & 1.00 {\scriptsize$\pm$0.00} \\
Lexical  & Control & DSGC           & 1.00 {\scriptsize$\pm$0.00} \\
Lexical  & Target  & Sim.-only & 0.03 {\scriptsize$\pm$0.18} \\
Lexical  & Target  & No-graph DSGC  & 0.03 {\scriptsize$\pm$0.18} \\
Lexical  & Target  & DSGC           & \textbf{0.90} {\scriptsize$\pm$0.31} \\
\midrule
Sentence & Control & Sim.-only & 1.00 {\scriptsize$\pm$0.00} \\
Sentence & Control & No-graph DSGC  & 1.00 {\scriptsize$\pm$0.00} \\
Sentence & Control & DSGC           & 1.00 {\scriptsize$\pm$0.00} \\
Sentence & Target  & Sim.-only & 0.23 {\scriptsize$\pm$0.43} \\
Sentence & Target  & No-graph DSGC  & 0.23 {\scriptsize$\pm$0.43} \\
Sentence & Target  & DSGC           & \textbf{1.00} {\scriptsize$\pm$0.00} \\
\bottomrule
\end{tabular}
\caption{Regime-level full-chain retention in the fixed main evaluation
suite (mean $\pm$ std over $15$~seeds). \texttt{Sim.-only} denotes the
standard similarity-only retention baseline. Bold values indicate
target-chain retention under one-hop propagation.}
\label{tab:main}
\end{table}

The control-target contrast in Table~\ref{tab:main} delivers a
coherent picture across both encoders. All three methods achieve
perfect retention ($1.00$) in the control regime: because the necessary
chain blocks share sufficient surface evidence with the query, standard
similarity-based retention easily preserves them. This saturation
indicates that the controls are retrieval-friendly: when chain blocks
are query-aligned, similarity-based retention is sufficient, and DSGC
does not degrade it.

In the target regime, the divergence between semantic similarity and
structural necessity becomes visible. Under the strict lexical encoder,
both baselines retain the full chain in only $0.03$ of runs, showing
the vulnerability of indirect prerequisites under budget pressure. DSGC mitigates this failure, preserving the full reasoning
chain in the vast majority of runs ($0.90$). The same pattern holds
under the denser sentence encoder, where DSGC reaches perfect retention
($1.00$) while the similarity baseline still frequently fails ($0.23$).

The exact empirical equality between the similarity-only and no-graph
DSGC baselines is a mathematical consequence of our controlled
implementation: setting $\lambda_\pi=0$ collapses the DSGC score to
raw semantic relevance ($I_i=r_i$). Because $\exp$ is monotone, this
ranking identity is exact, attributing the observed ranking difference
to the one-hop dependency propagation term.

\begin{table}[H]
\centering
\small
\begin{tabular}{@{}llll@{}}
\toprule
Encoder & Template & Sim.-only & DSGC \\
\midrule
Lexical  & \texttt{role}    & 0.00 {\scriptsize$\pm$0.00} & 1.00 {\scriptsize$\pm$0.00} \\
Lexical  & \texttt{profile} & 0.07 {\scriptsize$\pm$0.26} & 0.80 {\scriptsize$\pm$0.41} \\
Sentence & \texttt{role}    & 0.13 {\scriptsize$\pm$0.35} & 1.00 {\scriptsize$\pm$0.00} \\
Sentence & \texttt{profile} & 0.33 {\scriptsize$\pm$0.49} & 1.00 {\scriptsize$\pm$0.00} \\
\bottomrule
\end{tabular}
\caption{Granular breakdown of target template performance. The Profile
Target is more difficult than the Role Target under the lexical encoder,
consistent with the one-hop limitation illustrated in
Figure~\ref{fig:targets}. No-graph DSGC performance identically matches
the similarity-only baseline and is therefore omitted.}
\label{tab:templates}
\end{table}

\paragraph{Resolving the target regime.} Table~\ref{tab:templates}
breaks down target performance. Under the lexical encoder, the Profile
Target is the only setting where DSGC still fails on some seeds ($0.80$),
matching the predicted one-hop limit: DSGC rescues the immediate
prerequisite $c_2$ from query-facing $c_3$, but the oldest prerequisite
$c_1$ can remain vulnerable when it has too little residual lexical
support. Under the denser sentence encoder, $c_1$ receives enough
semantic support for DSGC to reach perfect retention. The next section
tests this boundary at the trace level.

\section{Trace Diagnostics}
\label{sec:traces}

Per-seed traces test a narrow diagnostic prediction: do the residual
lexical Profile failures preserve the rescued pair $(c_2,c_3)$ while
losing only the oldest prerequisite $c_1$?

\begin{table}[H]
\centering
\footnotesize
\setlength{\tabcolsep}{4pt}
\begin{tabular}{@{}ccccc@{}}
\toprule
Seed & Ranks $(c_1,c_2,c_3)$ & Top non-chain & Displ. & Margin \\
\midrule
3  & $(10,1,2)$ & \texttt{h1} & $c_1$ & 0.043 \\
8  & $(6,1,2)$  & \texttt{f9} & $c_1$ & 0.007 \\
14 & $(6,1,2)$  & \texttt{f3} & $c_1$ & 0.043 \\
\bottomrule
\end{tabular}
\caption{Diagnostic breakdown of Profile Target failures under the
lexical encoder. Prefix \texttt{h} denotes a honeypot distractor;
\texttt{f} denotes a background filler. In all failing instances, the
displaced block is exclusively the multi-hop prerequisite $c_1$.
The Margin column reports the final retention score differential
between the lowest-retained competitor and $c_1$; consistently narrow
values ($<0.05$) indicate marginal budget-cutoff displacements rather
than wholesale ranking collapses.}
\label{tab:failures}
\end{table}

\paragraph{Trace evidence for the one-hop boundary.}
Our evaluation framework logs per-seed ranking and displacement traces.
If the one-hop account is correct, residual DSGC failures should
preserve $c_2$ and $c_3$ while displacing $c_1$: the immediate
prerequisite ($c_2$) is rescued by propagation from the query-facing
block ($c_3$), but the structurally distant prerequisite ($c_1$) is
not. Table~\ref{tab:failures} matches this pattern. In all three
failing instances of the lexical Profile Target, $c_2$ and $c_3$
occupy the top two ranks; however, the oldest prerequisite $c_1$
drops to rank $6$ or $10$ and is subsequently evicted by a non-chain
distractor. In the remaining $12$ seeds, $c_1$ retains sufficient
residual lexical support to clear the budget cutoff, indicating that
the failure is marginal rather than systematic.

\paragraph{Anatomy of the residual failures.} Each failure decomposes
into a displaced chain block, an evicting non-chain competitor, and a
narrow score margin: in all three cases $c_1$ is edged out by a
non-chain block---honeypot \texttt{h1} or fillers \texttt{f9},
\texttt{f3}---with minimal lexical overlap, and the margins stay below
$0.05$, indicating marginal cutoff displacements rather than wholesale
ranking collapses. Seed~8 is informative because $c_1$ is displaced by
a plain background filler rather than a honeypot: the residual failure
is not limited to adversarial distractors; ordinary filler blocks can
also cross the cutoff.

\paragraph{Generating a falsifiable hypothesis.} These traces define a
concrete next test for $k$-hop propagation: a variant with $k\geq 2$
should recover these exact three seeds without degrading the perfect
retention of the controls and the Role Target. The narrowness of the
observed margins suggests that lightweight structural heuristics---such
as bounded neighborhood expansions---may suffice to close this gap
without the cost of deep multi-hop diffusion.

\section{Robustness and Failure Boundaries}
\label{sec:robust}

The trace evidence establishes the mechanistic picture; we now test
whether the same mechanism holds across propagation weights and budget
pressure.

\begin{table}[H]
\centering
\small
\begin{tabular}{@{}llll@{}}
\toprule
Template            & Encoder  & $\lambda_\pi$ & Retention \\
\midrule
\texttt{role}       & Lexical  & 0.5 & 1.00 {\scriptsize$\pm$0.00} \\
\texttt{role}       & Lexical  & 1.0 & 1.00 {\scriptsize$\pm$0.00} \\
\texttt{role}       & Lexical  & 2.0 & 1.00 {\scriptsize$\pm$0.00} \\
\texttt{profile}    & Lexical  & 0.5 & 0.67 {\scriptsize$\pm$0.49} \\
\texttt{profile}    & Lexical  & 1.0 & 0.80 {\scriptsize$\pm$0.41} \\
\texttt{profile}    & Lexical  & 2.0 & 0.87 {\scriptsize$\pm$0.35} \\
\texttt{role}       & Sentence & 0.5--2.0 & 1.00 {\scriptsize$\pm$0.00} \\
\texttt{profile}    & Sentence & 0.5--2.0 & 1.00 {\scriptsize$\pm$0.00} \\
\bottomrule
\end{tabular}
\caption{Hyperparameter stability on target templates. Mean retention
($\pm$ std over $15$~seeds) across varying propagation weights
($\lambda_\pi$). The results indicate that performance is not sensitive
to precise tuning within the tested range.}
\label{tab:lambda}
\end{table}
\vspace{-1.8em}

\paragraph{Hyperparameter stability ($\lambda_\pi$).} We evaluate the
propagation weight across $\lambda_\pi\in\{0.5,1.0,2.0\}$ to verify
local sensitivity around the main-suite default ($\lambda_\pi=1.0$).
As shown in Table~\ref{tab:lambda}, structural gains remain consistently
positive across this range: the Role Target is fully saturated under the
lexical encoder, the Profile Target improves monotonically, and the
sentence encoder achieves perfect retention at all tested values. These
results indicate that the gain is not sensitive to the exact value of
$\lambda_\pi$ in the tested range.

\begin{table}[H]
\centering
\small
\begin{tabular}{@{}llll@{}}
\toprule
Template & Mult. & Sim.-only & DSGC \\
\midrule
\texttt{role}    & 1.0 & 0.00 {\scriptsize$\pm$0.00} & 0.40 {\scriptsize$\pm$0.51} \\
\texttt{role}    & 2.0 & 0.00 {\scriptsize$\pm$0.00} & 1.00 {\scriptsize$\pm$0.00} \\
\texttt{role}    & 5.0 & 0.27 {\scriptsize$\pm$0.46} & 1.00 {\scriptsize$\pm$0.00} \\
\texttt{profile} & 1.0 & 0.00 {\scriptsize$\pm$0.00} & 0.53 {\scriptsize$\pm$0.52} \\
\texttt{profile} & 2.0 & 0.07 {\scriptsize$\pm$0.26} & 0.80 {\scriptsize$\pm$0.41} \\
\texttt{profile} & 5.0 & 0.60 {\scriptsize$\pm$0.51} & 1.00 {\scriptsize$\pm$0.00} \\
\bottomrule
\end{tabular}
\caption{Budget pressure dynamics under the lexical encoder. Full-chain
retention (mean $\pm$ std over $15$~seeds) across varying budget
multipliers ($M$), where
$C=\lceil M\sum_{b_i\in\mathrm{chain}(q)}s_i\rceil$. DSGC separates most
clearly from similarity-only retention in the intermediate regime
($M=2.0$), where similarity-based retention fails.}
\label{tab:budget}
\end{table}
\vspace{-1.8em}

\paragraph{Budget pressure dynamics.} Table~\ref{tab:budget} maps
retention behavior across three budget regimes:
\begin{itemize}[leftmargin=1.3em,itemsep=1pt,topsep=2pt]
  \item \textbf{Tight} ($M=1.0$): Both policies struggle, as the
    minimal budget cannot comfortably encompass the full reasoning chain
    alongside competitive distractors.
  \item \textbf{Intermediate} ($M=2.0$): The main separation regime.
    DSGC retains the structurally necessary chain while similarity-only
    retention does not.
  \item \textbf{Loose} ($M=5.0$): The performance gap narrows because
    the budget is large enough to passively retain nearly the entire
    context store.
\end{itemize}
Notably, the similarity-only curve is not strictly monotonic on the
Role Target, indicating that larger budgets do not automatically repair
structurally misranked prerequisites when competitive distractors remain
near the cutoff. This pattern suggests
that dependency-aware retention is most useful when the budget is large
enough to hold the chain but small enough for distractors to compete at
the cutoff.

\paragraph{Sliding window baseline.} We also compare against a
recency-based sliding-window policy. This baseline retains the full
chain in only $0.07$ of runs across all four templates under the
lexical encoder at $M=2.0$, failing even on the elementary control
scenarios. In this benchmark, recency is therefore neither a reliable
proxy for relevance nor a sufficient mechanism for preserving logical
prerequisite chains. The main comparison focuses on structured retention
policies, since the diagnostic question concerns how plausible memory
policies rank indirect prerequisites under budget pressure.

\begin{table}[H]
\centering
\small
\begin{tabular}{@{}llll@{}}
\toprule
Template      & Encoder  & 20 blocks & 50 blocks \\
\midrule
\texttt{role}    & Lexical  & 1.00 {\scriptsize$\pm$0.00} & 0.73 {\scriptsize$\pm$0.46} \\
\texttt{profile} & Lexical  & 0.80 {\scriptsize$\pm$0.41} & 0.33 {\scriptsize$\pm$0.49} \\
\texttt{role}    & Sentence & 1.00 {\scriptsize$\pm$0.00} & 1.00 {\scriptsize$\pm$0.00} \\
\texttt{profile} & Sentence & 1.00 {\scriptsize$\pm$0.00} & 1.00 {\scriptsize$\pm$0.00} \\
\bottomrule
\end{tabular}
\caption{Context scaling limitations. Full-chain retention for DSGC at
$20$ versus $50$ total context blocks (mean $\pm$ std over $15$~seeds).
The data shows material degradation under the sparse lexical encoder
as context size increases, while the dense sentence encoder remains
saturated in this setting.}
\label{tab:scale}
\end{table}
\vspace{-1.8em}

\paragraph{Context scaling and negative results.} Context scaling
reveals the clearest boundary of the current one-hop rule. As shown in
Table~\ref{tab:scale}, when expanding from $20$ to $50$ total blocks,
DSGC under the dense sentence encoder remains perfectly saturated
($1.00$). However, performance under the sparse lexical encoder
degrades materially: retention on the Role Target falls from $1.00$ to
$0.73$, and the Profile Target drops from $0.80$ to $0.33$.

A likely mechanism is score diffusion: as the candidate pool expands,
the baseline lexical relevance mass spreads across a broader array of
distractors, forcing the fixed one-hop propagation signal into tighter
competition against a much larger field of weakly similar background
blocks at the budget cutoff. In these $50$-block tests, the one-hop
rule remains saturated with dense embeddings but degrades under sparse
lexical encoding.

\section{Discussion}
\label{sec:discussion}

\paragraph{What this establishes.} Structurally indirect prerequisite
eviction is a well-defined retention-stage failure mode. In this
benchmark, it is reproducible across seeds, visible in per-seed traces,
and mitigated by one-hop dependency propagation. Control saturation
indicates that the gain is targeted rather than a blanket score
inflation, while residual failures mark the expected beyond-one-hop
boundary.

\paragraph{Boundaries and negative results.} By using synthetic
templates with supplied dependency edges, the benchmark isolates
retention from graph induction, multi-step planning, and tool use.
The $50$-block lexical result is an important negative case: it shows
where the current one-hop rule breaks. Failure-mode analysis is most
useful when it marks both the operating regime and the failure
boundary.

\paragraph{Error modes under imperfect graphs.} The score
$I_i = r_i + \lambda_\pi \pi_i$ degrades asymmetrically under graph
errors. A missing prerequisite edge removes propagation support from one
block, reducing DSGC to the baseline similarity ranking for that block. A spurious edge instead elevates an unrelated block
through $\pi_i$, where it can displace a true chain block at the budget
cutoff. This asymmetry suggests a design preference for high-precision
graph induction: missing edges remove support, whereas spurious edges
can actively elevate unrelated blocks.

\paragraph{End-to-end answer evaluation.} The deterministic template
answer accuracy in our suite is by construction a function of full-chain
retention: when every $b_i\in\mathrm{chain}(q)$ survives, the templated
answer is recoverable from $S$, which is why we report retention as the
primary metric. End-to-end evaluation with an LLM reasoner tests a
separate axis: whether a model can use a complete retained chain
correctly, rather than whether the chain survives eviction at all.

\paragraph{The multi-hop extension.} The trace diagnostics suggest a
$k$-hop extension to close the residual failures in
Table~\ref{tab:failures}. Moving to $k>1$ pushes propagation toward
$\mathcal{O}(k|E|)$ and introduces score-diffusion artifacts that may
require damping or normalization. These choices affect the algorithmic
trade-off rather than only implementation complexity. Multi-hop
variants will need to balance structural reasoning depth against
production latency budgets and, in realistic agent settings, eventually
connect to future graph induction pipelines built from tool-use or
reasoning traces \cite{generative_agents2023,reflexion2023}.

\paragraph{Systems profile.} The one-hop restriction also preserves
a favorable systems profile: $r_i$ is a dense query--block inner
product, $\pi_i$ is a single sparse pass implementable as SpMV or
scatter-add, and retention reduces to score ordering plus a top-$k$
budgeted selection step, yielding $\mathcal{O}(Nd + |E| + N\log N)$
total cost. When the graph is sparse, the structural overhead follows
declared prerequisite edges rather than an $\mathcal{O}(N^2)$
all-pairs interaction pattern, leaving production latency measurements
to a separate systems evaluation. Graph induction can also be amortized
outside the per-step retention path: a tool-call trace in which a
lookup result gates a downstream action directly records a prerequisite
edge that a lightweight parser can index at ingestion and reuse across
later retention cycles.

\section{Conclusion}

We isolated a retention-stage failure mode---the eviction of
structurally indirect prerequisites---and provided an operational
definition, a fixed reproducible trigger, trace-level diagnostics, and
a minimal tested mitigation. Together, these results support a broader
principle: before retrieval can succeed, retention must keep the
reasoning chain live.

More broadly, agentic memory is better understood as a dynamic managed
system than as a static vector store. The operational question shifts
from ``Which block is most similar to the query?'' to ``Which blocks
must remain live for the reasoning chain to stay intact?'' DSGC
instantiates this idea in a one-hop, systems-friendly form, showing
that following prerequisite edges can substantially close the gap
between surface-similarity retention and chain-intact reasoning.

\paragraph{Reproducibility.} The public release at
\url{https://github.com/smkgenesis/dsgc} contains the full
artifact trail---protocol, postmortem, calibration, main suite, and
robustness pack---with a \texttt{reproduce.sh} entrypoint.

\paragraph{Use of AI assistance.} The author used large language model
assistants during code implementation and manuscript preparation. All
research design, analyses, and conclusions are the author's
responsibility.

\bibliographystyle{icml2026}
\bibliography{references}

\end{document}